\documentclass{article}
\usepackage{graphicx,amsmath,bm}

\title{Statistical Mechanics of Node-perturbation Learning with Noisy Baseline}

\author{\textsc{Kazuyuki Hara}$^{1}$\thanks{E-mail: hara.kazuyuki@nihon-u.ac.jp}, \textsc{Kentaro Katahira}$^{2}$\thanks{E-mail: katahira@mns.k.u-tokyo.ac.jp},  and \textsc{Masato Okada}$^{3,4}$\thanks{E-mail: okada@k.u-tokyo.ac.jp}}
\date{$^{1}$College of Industrial Technology, Nihon University, \\
1-2-1 Izumi-cho, Narashino, Chiba 275-8575, Japan \\
$^{2}$Graduate School of Environmental Studies, Nagoya University, \\
Furo-cho, Chikusa-ku, Nagoya, 464-8601 Japan\\
$^{3}$Graduate School of Frontier Sciences, The University of Tokyo,\\
5-1-5 Kashiwanoha, Kashiwa-shi, Chiba 277-8561, Japan\\
$^{4}$Riken Brain Science Institute, \\2-1 Hirosawa, Wako, Saitama 351-0198, Japan
}

\begin{document}
\maketitle

\begin{abstract}
Node--perturbation learning is a type of statistical gradient descent algorithm that can be applied to problems where the objective function is not explicitly formulated, including reinforcement learning. It estimates the gradient of an objective function by using the change in the object function in response to the perturbation. The value of the objective function for an unperturbed output is called a baseline. 
Cho et al. proposed node--perturbation learning with a noisy baseline. In this paper, we report on building the statistical mechanics of Cho's model and on deriving coupled differential equations of order parameters that depict learning dynamics. We also show how to derive the generalization error by solving the differential equations of order parameters. On the basis of the results, we show that Cho's results are also apply in general cases and show some general performances of Cho's model.
\end{abstract}

\section{Introduction}
Learning in neural networks\cite{Widrow1990} can be formulated as an optimization of an objective function that quantifies the system's performance. This is achieved by following the gradient of the objective function with respect to the tunable parameters of the system. This optimization is computed directly by calculating the gradient explicitly and by updating the parameters using a small step in the direction of the locally greatest improvement. However, computing a direct gradient to follow can be problematic. For instance, reinforcement learning has no explicit form of objective function, so we cannot calculate the gradient.

A stochastic gradient--following method to estimate gradient information has been proposed for problems where a true gradient is not directly given. Node--perturbation learning (NP learning)\cite{williams1992} is one of the stochastic learning algorithms. NP learning estimates the gradient of an objective function by using the change in the objective function in response to a small perturbation. NP learning can be formulated as reinforcement learning with a scalar reward, and all the weight vectors are updated by using the scalar reward, while the gradient method uses the target vector. Hence, as well as being useful as a neural network learning algorithm, NP learning can be formulated as reinforcement learning \cite{williams1992,Sprekeler2009}, or it can be used in a brain model \cite{Fiete2006,Fiete2007}. 

The learning curve of linear NP learning was derived by Werfel et al.\cite{werfel2005} by calculating the average of the squared error over all possible inputs. They formulated the NP  learning as online learning. We were inspired by Werfel's formulation of linear NP  learning and analyzed the behavior of generalization by using the statistical mechanics method\cite{KroghHertz1992,Hara2010}. A compact description of the learning dynamics can be obtained by using statistical mechanics, which assumes a large limit of the system size $N$ and provides an accurate model of mean behavior for a realistic $N$\cite{nishimori2001,Engelbook2001,Krogh1992}. We derived the order parameter equations that describe the dynamics of the system at the thermodynamic limit of $N \rightarrow \infty$ and solved the dynamics of the generalization error of linear NP  learning. Our results\cite{Hara2010} are independent of the system size, so general performance can be shown; however, Werfel's results depend on the system size. Moreover, our results have order parameter equations and a generalization error, so we can analyze the dynamics of the learning behavior. Moreover, Werfel's results only have a learning curve.

Noise is present all the time in real circumstances. Cho et al. proposed NP  learning with a noisy baseline\cite{Cho2011}. In their paper, they analyzed the behavior of the learning curve of their model by averaging the squared error over all possible inputs. They showed that the residual error of the learning curve becomes the smallest when the ratio of the variance of baseline noise to perturbation noises is one. 

In this paper, we report on building the statistical mechanics of Cho's model based on online learning\cite{Saadbook,Biehl1995}. We follow Cho's formulation, we formulate the learning equation of NP learning with a noisy baseline, and then derive order parameter equations of the learning system by using the statistical mechanics method. The order parameters depict the dynamics of the learning, and order parameter equations are derived assuming the thermodynamic limit of system size $N \rightarrow \infty$. The objective of building the statistical mechanics of Cho's model is to obtain the generalization error of the model. We formulated the order parameter equations \cite{hara2005,Biehl1994,Reents1998} and then solved the analytical solutions of the order parameter equations. The generalization error is a function of the order parameters, so we solved the time dependence of the generalization error by using the solutions of the order parameters. On the basis of the results, we show that Cho's results are also apply in the general case and show the general performance of Cho's model.

The remainder of this paper is organized as follows. In the next section, we first introduce the learning rule with a noiseless baseline and then show the rule with a noisy baseline. Next, we show some order parameter equations of Cho's model. 
In Sec. 3, we solve closed order parameter equations analytically and give an analytical solution of the generalization error. We then compare the numerical solution of the generalization error with that of simulation results to show the validity of the analytical solutions. We also discuss the optimization of the generalization error based on fast relaxation or minimum residual error. The final section concludes with given results.

\section{Formulation}
In this section, we describe the formulation of the teacher and student networks and the derivation of a learning rule of the NP learning algorithm using circumstance noise (baseline noise) and perturbation noise through a linear perceptron. 

\subsection{Model}
\label{model}
Here, we formulate the teacher and student networks and an NP learning algorithm utilizing a teacher--student formulation. In the teacher--student formulation, the teacher network generates a target of the student network for a given input. By introducing the teacher, we can directly measure the similarity of the student weight vector to the teacher weight vector. 

\begin{figure}[b] 
\begin{center} 
\includegraphics[width=12cm]{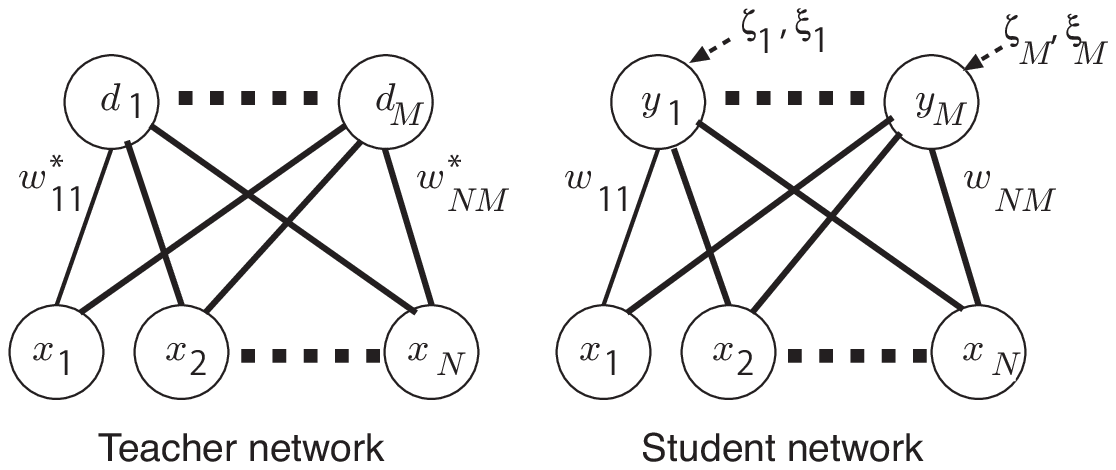} 
\end{center} 
\caption{\label{networks} Structure of teacher and student networks. Both networks have the same structure. Two independent noises, $\xi_k$ and $\zeta_k$, are used.} 
\end{figure}

We assume that the teacher and student networks receive $N$-dimensional input $\bm{x}^{(m)}=(x_1^{(m)}, \ldots , x_N^{(m)})$ at the $m$th learning iteration as shown in Fig. \ref{networks}\cite{werfel2005}. Here, we assume the existence of a teacher network vector $\bm{w}_k^{\ast}$ that produces a desired output, so the teacher output $d_k$ is the target of the student output $y_k$. The learning iteration $m$ is ignored in the figure. We assume that the elements $x_i^{(m)}$ of independently drawn input $\bm{x}^{(m)}$ are uncorrelated random variables with zero mean and unit variance; that is, the $i$th element of the input is drawn from an identical Gaussian distribution $P(x_i)$. In this paper, the thermodynamic limit of $N \rightarrow \infty$ is assumed. In the thermodynamic limit, the law of large numbers and the central limit theorem can apply. We can then depict the system behavior by using a small number of parameters. In this limit, $|| \bm{x}^{(m)} ||=\sqrt{N}$ is satisfied. Here, $|| \cdot ||$ denotes the norm of a vector.

The teacher network shown in Fig. \ref{networks} has $N$ inputs and $M$ outputs and are identical to $M$ linear perceptrons (teachers). Teachers are not subject to learning. Therefore, the weight matrix of teacher networks with $M \times N$ elements $\bm{w}^{\ast}$ is fixed throughout the learning. The teacher weight vectors $\{\bm{w}^{\ast}_k\}$, $\bm{w}^{\ast}_k=(w^{\ast}_{1k}, \ldots , w^{\ast}_{Nk})$ are $N$-dimensional vectors, and each element $w^{\ast}_{jk}, \ j=1, \ldots , N$ of the teacher weight vectors $\bm{w}^{\ast}_k$ is drawn from an identical Gaussian distribution of zero mean and variance $1/N$. Assuming the thermodynamic limit of $N \rightarrow \infty$, the norm of the $k$th teacher weight vector is $\| \bm{w}_k^{\ast}\|=1$. The output of the $k$th teacher $d_k^{(m)}$ for $N$-dimensional input $\bm{x}^{(m)}$ at the $m$th learning iteration is 

\begin{equation}
d_k^{(m)}=\bm{w}^{\ast}_k \cdot \bm{x}^{(m)}=\sum_{j=1}^N w^{\ast}_{jk} x_j^{(m)}.
\label{d}
\end{equation}

\noindent 
The distribution of the $k$th output of the teacher networks follows a Gaussian distribution of zero mean and unit variance at the thermodynamic limit. 

Student network has $N$ inputs and $M$ outputs and are identical to $M$ linear perceptrons (students). Students are subject to learning. For the sake of analysis, we assume that each element of the initial student weight $w_{jk}^{(0)}$ is drawn from an identical Gaussian distribution of average zero and variance $1/N$. Assuming the thermodynamic limit of $N \rightarrow \infty$, the norm of the $k$th initial student weight vector is $\| \bm{w}_k^{(0)}\|=1$. The output of the $k$th student $y_k^{(m)}$ for $N$-dimensional input $\bm{x}^{(m)}$ at the $m$th learning iteration is 

\begin{equation}
y_k^{(m)}=\bm{w}_k^{(m)} \cdot \bm{x}^{(m)}=\sum_{j=1}^N w_{jk}^{(m)} x_j^{(m)}.
\label{y}
\end{equation}

\noindent 
At the thermodynamic limit, the distribution of $y_k$ obeys the Gaussian distribution of the mean zero and variance $Q_{kk}$. Here, $Q_{kk}$ denotes $Q_{kk} = \bm{w}_k \cdot \bm{w}_k$. 

We utilize the squared error as an error function. Following Werfel et al. \cite{werfel2005}, the squared error $E^{(m)}$ at the $m$th iteration is defined as

\begin{equation}
E^{(m)}=\|\bm{d}-\bm{y}\|^2=\frac{1}{2}\sum_{k=1}^M (d_k^{(m)}-y_k^{(m)})^2. 
\label{squared_error} 
\end{equation}

Next, we formulate the NP learning\cite{werfel2005}. The objective function in our learning system is the squared error $E^{(m)}$. It estimates the gradient of the squared error by using the change in the squared error in response to perturbed student output and to unperturbed student output. The squared error for perturbed student output, $E_{\xi}^{(m)}$, is defined as 

\begin{equation}
E_{\xi}^{(m)}=\|\bm{d}-(\bm{y}+\bm{\xi})\|^2=\frac{1}{2}\sum_{k=1}^M [ d_k^{(m)}-(y_k^{(m)}+\xi_k^{(m)}) ]^2.
\label{squared_error_noise2} 
\end{equation}

\noindent
If the addition of the noise vector $\bm{\xi}$ lowers the error, the student weight vectors are adjusted in the direction of the noise vector. Here, each element $\xi_k$ of the noise vector $\bm{\xi}$ is drawn from a Gaussian distribution of zero mean and variance $\sigma_{\xi}^2$ . Werfel et al. proposed the learning equation\cite{werfel2005} as

\begin{equation}
w_{jk}^{(m+1)}=w_{jk}^{(m)}-\frac{\eta}{N\sigma_{\xi}^2}(E_{\xi}^{(m)}-E^{(m)})\xi_k^{(m)} x_j^{(m)}. 
\label{werfel} 
\end{equation}

\noindent 
Here, $E_{\xi}^{(m)}-E^{(m)}$ is the difference between the error with and without noise. Note that the difference $E_{\xi}^{(m)}-E^{(m)}$ is assigned for each output $y_k^{(m)}$ from the independence of noise $\xi_k^{(m)}$ for each output. $\eta$ denotes the learning rate. NP learning defined by Eq. (\ref{werfel}) is referred to as single NP (SNP) in this paper.

Next, we follow Cho's model \cite{Cho2011}, which uses two independent noises. We use a noisy baseline, so Eq. (\ref{squared_error}) is replaced by

\begin{equation}
E_{\zeta}^{(m)}=\|\bm{d}-(\bm{y}+\bm{\zeta})\|^2=\frac{1}{2}\sum_{k=1}^M [ d_k^{(m)}-(y_k^{(m)}+\zeta_k^{(m)}) ]^2.
\label{squared_error_noise1} 
\end{equation}

\noindent
Here, each element of $\zeta_k$ of the baseline noise vector $\bm{\zeta}$ is drawn from the Gaussian distribution of zero mean and variance $\sigma_{\zeta}^2$. We assume that the noises $\xi_k$ and $\zeta_k$ are independent of each other. Then, the learning equation of Cho's model\cite{Cho2011} is given as

\begin{equation}
w_{jk}^{(m+1)}=w_{jk}^{(m)}-\frac{\eta}{N\sigma_{\xi}^2}(E_{\xi}^{(m)}-E_{\zeta}^{(m)})\xi_k^{(m)} x_j^{(m)}. 
\label{double} 
\end{equation}

\noindent
When $\zeta_k=0$, the proposed method is identical to the SNP. NP learning using Eq. (\ref{double}) is called double NP (DNP) in this paper. 

\subsection{Theory}
\label{theory}
In this paper, we describe our consideration of the thermodynamic limit of $N \rightarrow \infty$ to analyze the dynamics of the generalization error of our system through statistical mechanical methods. In the following, the subscript $k$ means that the term is related to the output, which is subject to learning, and the subscript $l$ means that the term is not related to the output.

By expanding Eq. (\ref{double}), the learning equation of DNP is written as

\begin{align}
w_{jk}^{(m+1)}&=w_{jk}^{(m)} -\frac{\eta}{N\sigma_{\xi}^2} (E_{\xi}^{(m)}-E_{\zeta}^{(m)} )\xi_k^{(m)} x_j^{(m)} \nonumber\\
&=w_{jk}^{(m)}+\frac{\eta}{N} \delta_k^{(m)} x_j^{(m)},\label{le-dnp}
\end{align}
\begin{align}
\delta_k^{(m)}&=\frac{1}{2\sigma_{\xi}^2}\Biggl\{ 2(\xi_k^2-\xi_k\zeta_k)(d_k-y_k)-\xi_k^3+\xi_k\zeta_k^2\Biggr.\nonumber\\
&\Biggl.+\sum_{l\neq k}\bigl[ 2(\xi_k\xi_l-\xi_k\zeta_l)(d_l-y_l)-\xi_k\xi_l^2+\xi_k\zeta_l^2\bigr]\Biggr\}.
\label{prop-delta}
\end{align}

\noindent 
In Eq. (\ref{prop-delta}), we omitted the learning iteration $m$ for simplicity. 

From Eq. (\ref{double}), if the addition of the noise vector $\bm{\xi}^{(m)}$ decreases the error, i.e., $E_{\xi}-E_{\zeta}<0$, the student weight vectors are added in the direction of the sign of $\xi^{(m)}_k x_j^{(m)}$. Note that the difference $(E_{\xi}^{(m)}-E_{\zeta}^{(m)})$ is assigned to each output $y_k^{(m)}$. However, from the independence of the noise $\xi_k^{(m)}$ from $\xi_l^{(m)}, l\neq k$, the learning progresses independently at each output.

The generalization error is defined as the squared error Eq. (\ref{squared_error}) averaged over possible inputs: 

\begin{align}
\varepsilon_g&=\int \mbox{d}\bm{x} P(\bm{x}) E\nonumber \\
&= \int \mbox{d}\bm{x}P(\bm{x}) \frac{1}{2}\sum_{k=1}^M \left(\sum_{i=1}^N w_{ik}^{\ast}x_i-\sum_{i=1}^N w_{ik} x_i\right)^2. 
\label{eg1} 
\end{align}

\noindent 
This calculation is the $N$th Gaussian integral with $\bm{x}$, and it is difficult to calculate. To overcome this difficulty, we utilize a coordinate transformation from $\bm{x}$ to $\{d_k\}$ and $\{y_k\}, k=1 \ldots M$, in Eqs.  (\ref{d}) and (\ref{y}), and we rewrite Eq. (\ref{eg1}) as

\begin{align}
\varepsilon_g&= \int \prod_{k=1}^M \mbox{d}d_k \mbox{d}y_k P(d_1,\ldots d_M, y_1, \ldots y_M)  \frac{1}{2}\sum_{k=1}^M \left(d_k-y_k\right)^2 \nonumber \\
&=\frac{1}{2}\sum_{k=1}^M \left(\left<d_k^2\right>-2\left<d_k y_k\right>+\left< y_k^2\right>\right) \nonumber \\
&=\sum_{k=1}^M \frac{1}{2}(T_{kk}-2R_{kk}+Q_{kk}).
\label{eg2} 
\end{align}

\noindent
Here, $\left< \cdot \right>$ denotes the average over possible inputs. 
In this equation, $Q_{kk}$ is defined as $\bm{w}_k\cdot \bm{w}_k$, $R_{kk}$ is defined as $\bm{w}_k^{\ast}\cdot \bm{w}_k$, and $T_{kk}$ is defined as $\bm{w}_k^{\ast}\cdot \bm{w}_k^{\ast}$. 

Next, we show how to derive coupled differential equations of the order parameters $Q_{kk}$, $Q_{kl}$, $R_{kk}$, and $R_{kl}$. $Q_{kl}$ and $R_{kl}$ are considered because when updating $\bm{w}_k$, $\delta_k$, which includes $\bm{w}_l$, is used. Then, $\bm{w}_k$ and $\bm{w}_l$ may be correlated. 
The differential equations for online learning\cite{Biehl1995} are

\begin{align}
\frac{\mbox{d}Q_{kk}}{\mbox{d}t}&=2\eta \left< \delta_k y_k \right> + \eta^2 \left< \delta_k^2\right>,\label{dl2dt} \\ 
\frac{\mbox{d}R_{kk}}{\mbox{d}t} &= \eta \left< \delta_k d_k\right>, \label{dr2dt}\\
\frac{\mbox{d}Q_{kl}}{\mbox{d}t}&=\eta ( \left< \delta_l y_k \right> + \left< \delta_k y_l\right> ) +\eta^2 \left< \delta_k \delta_l \right>,\label{dQ2dt} \\
\frac{\mbox{d}R_{kl}}{\mbox{d}t}&=\eta \left< \delta_l d_k \right>. \label{drdt}
\end{align}

\noindent 
Here, $\left<\cdot \right>$ means the average over all possible inputs. The time $t=m/N$ becomes continuous at the limit of $N \rightarrow \infty$. $\delta_k$ depends on the learning rule. For DNP, $\delta_k$ is denoted by Eq. (\ref{prop-delta}). We substitute Eq. (\ref{prop-delta}) into Eqs.  (\ref{dl2dt}) to (\ref{drdt}): 

\begin{align}
\frac{\mbox{d}Q_{kk}^2}{\mbox{d}t}&=2 \eta (R_{kk}-Q_{kk}) + \eta^2 \Bigl[(3+\gamma)(T_{kk}-2R_{kk}+Q_{kk}) \Bigr. \nonumber \\
&\Bigl. +(1+\gamma)\sum_{k\neq l}^M(T_{ll}-2R_{ll}+Q_{ll}) +\frac{\sigma_{\xi}^2}{4}(M+2)(M(1-\gamma)^2+4)\Bigr],\label{dQ2dt2}\\
\frac{\mbox{d}R_{kk}}{\mbox{d}t}&=\eta(T_{kk}-R_{kk}),\label{dRdt2}\\
\frac{\mbox{d}Q_{kl}^2}{\mbox{d}t}&=\eta(R_{kl}+R_{lk}-2Q_{kl})+2\eta^2(T_{kl}-R_{kl}-R_{lk}+Q_{kl}), \label{dQdt}\\
\frac{\mbox{d}R_{kl}}{\mbox{d}t}&=\eta(T_{kl}-R_{kl}). \label{dRdt}
\end{align}

\noindent
Here, we use $\left<\xi_k^2\right>=\left<\xi_l^2\right>=\sigma_{\xi}^2$, $\left<\xi_k^4\right>=\left<\xi_l^4\right>=3\sigma_{\xi}^4$, $\left< \xi_k^6\right>=15\sigma_{\xi}^6$, $\left<\xi_k\right>=\left<\xi_k^3\right>=\left<\xi_k^5\right>=0$. We defined $\gamma=\sigma_{\zeta}^2/\sigma_{\xi}^2$. 

Since $\bm{w}_k$ and $\bm{w}_l$ are updated with the same input $\bm{x}$, and $\delta_k$ includes $\xi_l$ as can be seen in Eq.  (\ref{prop-delta}), they may not be independent.
Equation (\ref{dQdt}) shows the correlation between the $k$th hidden weight vector and the $l$th hidden unit vector of the student. Equation (\ref{dRdt}) shows the overlap between the $k$th hidden weight vector of the teacher and the $l$th hidden weight vector of the student. Note that the effect of noise ($\xi_k$ and $\zeta_k$) appears only on the term of $\eta^2$ in Eq. (\ref{dQ2dt2}). The reason is that these noises have zero--mean Gaussian distributions. The cross--talk noise, which originates from the error of the other outputs, appears in Eq. (\ref{dQ2dt2}) from the average of the second--order cross--talk noise $\left< \xi_k^2 \xi_l^2 \right>$, while the average of the first--order $\left< \xi_k \xi_l \right>$ in Eqs.  (\ref{dQ2dt2}) and (\ref{dRdt}) is eliminated.

\section{Results}
In this section, we discuss the dynamics of the order parameters and their asymptotic properties and then show the derivation of the analytical solution of the generalization error. Finally, we discuss the validity of the analytical results in a comparison with simulation results.

For the sake of simplicity, the initial weight vectors of the teachers and students are homogeneously correlated, so we assume $Q_{kk}(0)=Q(0)$ and $R_{kk}(0)=R(0)$. From the symmetry of the evolution equation for updating the weight vector,

\begin{equation}
Q_{kk}(t) = Q(t), \ \ \ R_{kk}(t)= R(t), \label{conditions_Q_R}
\end{equation}
are obtained. 

Substituting Eq. (\ref{conditions_Q_R}) into Eqs. (\ref{dQ2dt2}) and (\ref{dRdt2}), we obtain

\begin{align}
\frac{\mbox{d}Q}{\mbox{d}t}&=2\eta(R-Q)+\eta^2\Bigl[(M(1+\gamma)+2)(1-2R+Q)+\frac{\sigma_{\xi}^2}{4}(M+2)(M(1-\gamma)^2+4)\Bigr], \label{dl3}\\
\frac{\mbox{d}R}{\mbox{d}t}&=\eta(1-R).
\label{dr3}
\end{align}

\noindent
Here, we use $T_{kk}=1$. Next, we solve the coupled differential equations of Eqs.  (\ref{dl3}), (\ref{dr3}), (\ref{dQdt}), and (\ref{dRdt}). Equation (\ref{dr3}) can be solved analytically as

\begin{equation}
R(t)=1-(1-R(0))\exp(-\eta t). \label{rt}
\end{equation}

\noindent
By substituting Eq.  (\ref{rt}) into Eq.  (\ref{dl3}) and solving it analytically, we obtain

\begin{align}
Q(t)&=\Biggl\{1-2R(0)+Q(0)-\frac{(M+2)[M(1-\gamma)^2+4]\eta\sigma_{\xi}^2}{4[2-(M(1+\gamma)+2)\eta]}\Biggr\}e^{-[2\eta-(M(1+\gamma)+2)\eta^2] t}\nonumber \\
& -2(1-R(0))e^{-\eta t}+\Bigl\{1+\frac{(M+2)[M(1-\gamma)^2+4]\eta\sigma^2}{4[2-(M(1+\gamma)+2)\eta]}\Bigr\}.
\label{lt}
\end{align}

We can also solve Eqs.  (\ref{dQdt}) and (\ref{dRdt}) in the same way as the aforementioned calculations. Equation (\ref{dRdt}) can be solved analytically as

\begin{equation}
R_{kl}(t)=T_{kl}-(T_{kl}-R_{kl}(0))\exp(-\eta t). \label{ana-rkl}
\end{equation}

\noindent
We also substitute Eq.  (\ref{ana-rkl}) into Eq.  (\ref{dQdt}) and solve it analytically, enabling us to obtain

\begin{align}
Q_{kl}(t)&=(T_{kl}-R_{kl}(0)-R_{lk}(0)+Q_{kl}(0))e^{-2(\eta-\eta^2)t}\nonumber \\
&-(2T_{kl}-R_{kl}(0)-R_{lk}(0)) e^{-\eta t}+T_{kl}\label{ana-Q}.
\end{align}

Since $\bm{w}_k$ and $\bm{w}_l$ are updated with the same input $\bm{x}$, and $\delta_k$ includes $\xi_l$ as can be seen in Eq.  (\ref{prop-delta}), they may not be independent.
From Eqs.  (\ref{ana-rkl}) and (\ref{ana-Q}), $R_{kl}$ and $Q_{kl}$ converge into $T_{kl}$ when $t \rightarrow \infty$. As shown in Sec. \ref{model}, each element $w^{\ast}_{ik}, \ i=1, \ldots , N$, of the teacher weight vectors $\bm{w}^{\ast}_k$ is drawn from an identical Gaussian distribution of zero mean and variance $1/N$. Then, $T_{kl}=0$ at the thermodynamic limit of $N \rightarrow \infty$. Therefore, $Q_{kl}$ and $R_{kl}$ vanish in this case. 

By using the assumptions of $R_{kk}(0)=R(0)$ and $Q_{kk}(0)=Q(0)$, the time course of the generalization error is given as

\begin{equation}
\varepsilon_g(t)=\frac{M}{2}(1-2R(t)+Q(t))=M\varepsilon_g^{\ast}(t).
\end{equation}

\noindent
Here, $\varepsilon_g^{\ast}(t)$ is the generalization of one output. Therefore, the generalization error $\varepsilon_g(t)$ is obtained by  substituting Eqs.  (\ref{lt}) and (\ref{rt}) into Eq.  (\ref{eg2}):

\begin{align}
\varepsilon_g(t)=&M\Biggl\{\Bigl(\varepsilon_g^{\ast}(0)-RE\Bigr) e^{-[2\eta-(M(1+\gamma)+2)\eta^2] t}+RE\Biggr\},\label{eg3}\\
RE&=\frac{(M+2)[M(1-\gamma)^2+4]\eta\sigma_{\xi}^2}{8[2-(M(1+\gamma)+2)\eta]}.\label{lerr}
\end{align}

\noindent
Here, $-[2\eta-(M(1+\gamma)+2)\eta^2]t$ is the relaxation speed of the generalization error, and $RE$ is the residual error. This equation shows that the residual error vanishes at the limit of $\eta \rightarrow 0$.

\begin{figure}[ht] 
\begin{center}
\includegraphics[width=8.5cm]{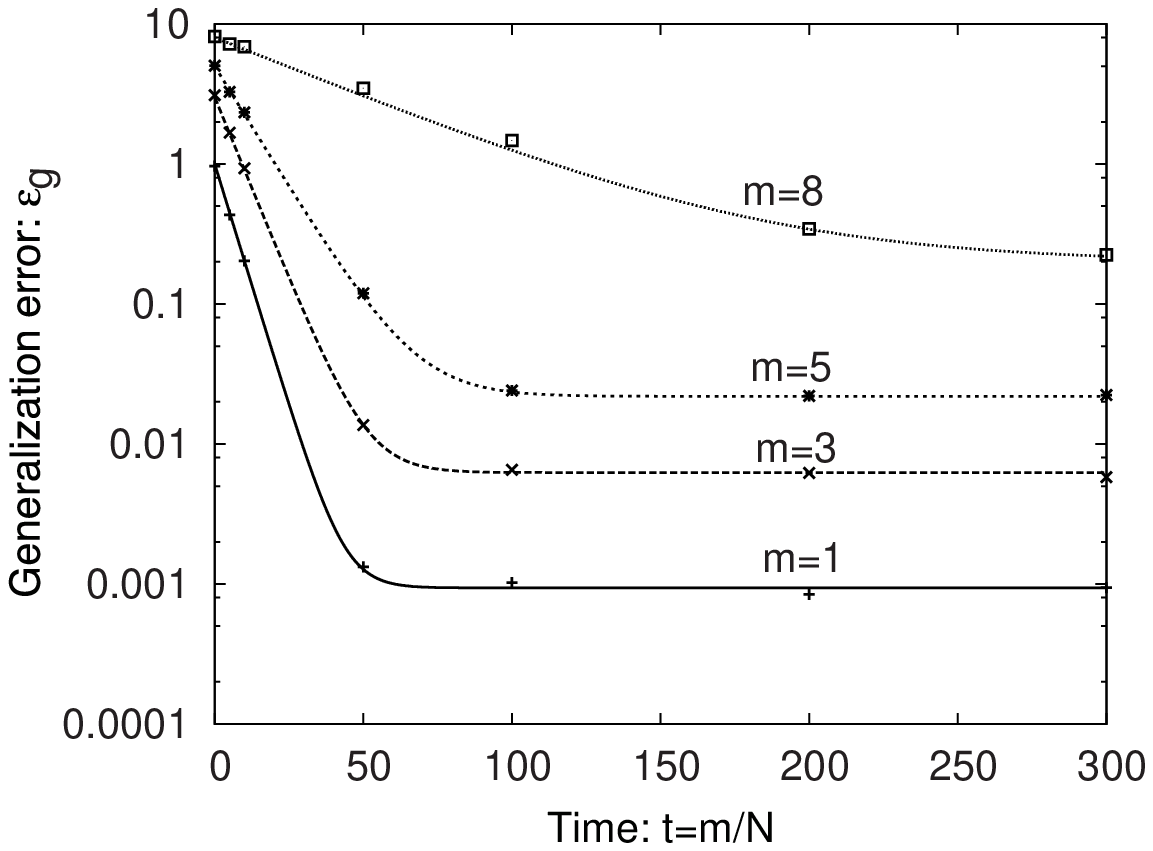} 
\end{center}
\caption{\label{fig:eg} Time course of the generalization error.} 
\end{figure}

Next, to show the validity of the analytical solutions, we compare the numerical calculation of analytical solutions with those of computer simulations. The time course of the generalization error is shown in Fig. \ref{fig:eg}. The number of outputs of the teacher and student is $M=1, 3, 5, 8$. We assume $Q_{kk}(0)=Q(0)$ and $R_{kk}(0)=R(0)$. The variances of the perturbation noise and baseline noise are set to $\sigma_{\xi}^2=\sigma_{\zeta}^2=0.01$. The learning step size is set to $\eta=0.1$. In the numerical calculation of the analytical solution, we set $T_{kk}=1$, $T_{kl}=0$, $Q(0)=1$, $Q_{kl}(0)=0$, $R(0)=0$, and $R_{kl}(0)=0$. In the computer simulation, $N=1000$. Each element of the teacher weight vectors and the initial student weight vectors are drawn from the Gaussian distribution of zero mean and variance $1/N$. Each element of the input vector is drawn from the Gaussian distribution of zero mean and unit variance.

In Fig. \ref{fig:eg}, analytical solutions for $M=1$ are drawn by the solid lines, those for $M=3$ are drawn by the chained lines, those for $M=5$ are drawn by the dotted lines, and those for $M=8$ are drawn by the broken lines. In the figure, the computer simulation results for $M=1$ are plotted by ``$+$," those for $M=3$ are plotted by ``$\times$," those for $M=5$ are plotted by ``$\ast$," and those for $M=8$ are plotted by ``□." The horizontal axis in this figure is the learning time $t=m/N$. The vertical axis is the generalization error $\varepsilon_g$. 

The figure shows that the numerical calculation of the analytical solutions agreed with those of the computer simulations, demonstrating the validity of the analytical solutions. From the figure, the generalization error increases as $M$ increases; however, the change in the generalization error is not proportional to $M$. This may be caused by crosstalk noise added from other output units. 

Next, we discuss two optimizations: one is optimization of the relaxation time, and the other is minimization of residual error. We first discuss the optimization.

From Eq. (\ref{eg3}), the optimum learning step size $\eta_{opt}$ to make the relaxation time the shortest is given by solving the condition $2(M(1+\gamma)+2)\eta-2=0$:

\begin{equation}
\eta_{opt}=\frac{1}{M(1+\gamma)+2}.\label{opt-eta}
\end{equation}

\noindent
Under this condition, the residual error $RE_{\eta_{opt}}$ is given by 

\begin{equation}
RE_{\eta_{opt}}=\frac{(M+2)[M(1-\gamma)^2+4]\sigma_{\xi}^2}{8[M(1+\gamma)+2]}.\label{eta_opt_RE}
\end{equation}

\noindent
Therefore, the optimum ratio $\gamma_{\eta_{opt}}$ that minimizes the residual error (refer to Eq. (\ref{eta_opt_RE})) is given by solving $\frac{\partial RE_{\eta_{opt}}}{\partial \gamma}=0$:

\begin{equation}
\gamma_{\eta_{opt}}=\frac{2\sqrt{M^2+3M+1}-(M+2)}{M}.\label{opt-eta-gamma}
\end{equation}

\begin{figure}[ht]
\includegraphics[width=8cm]{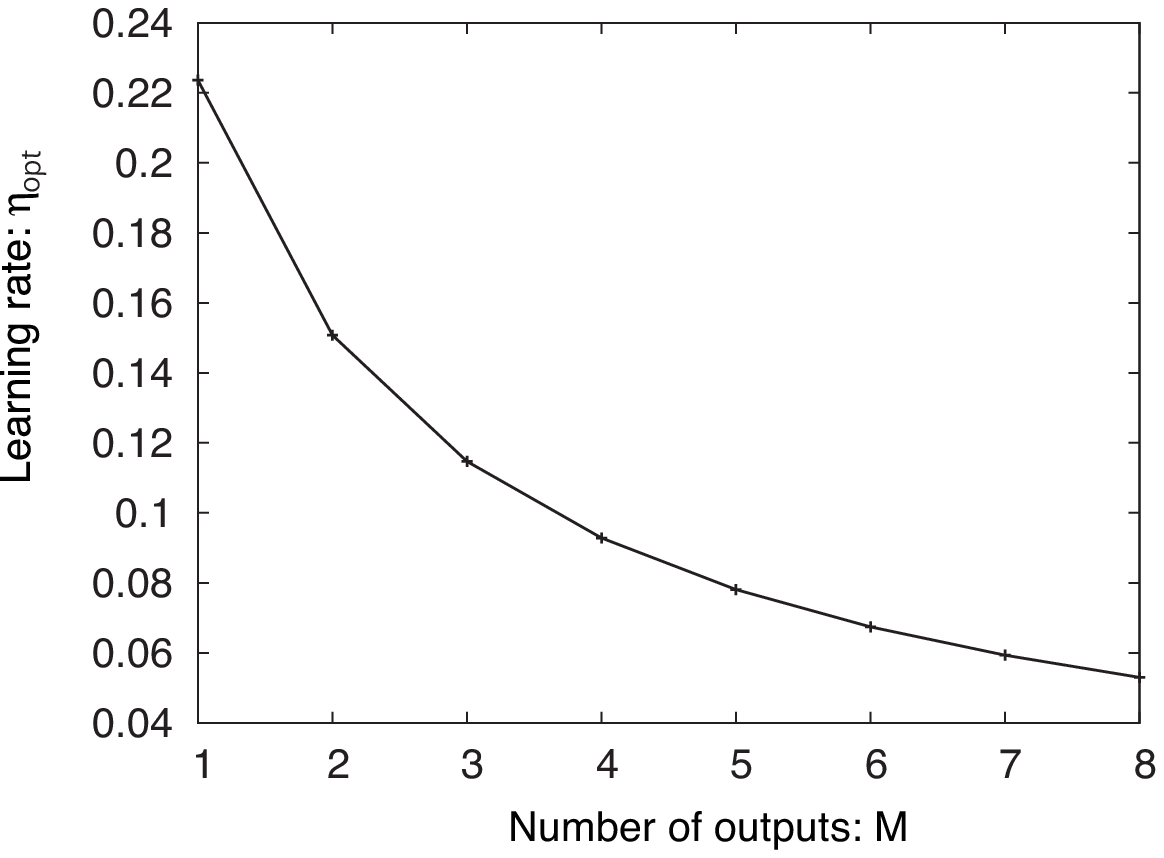}
\includegraphics[width=8cm]{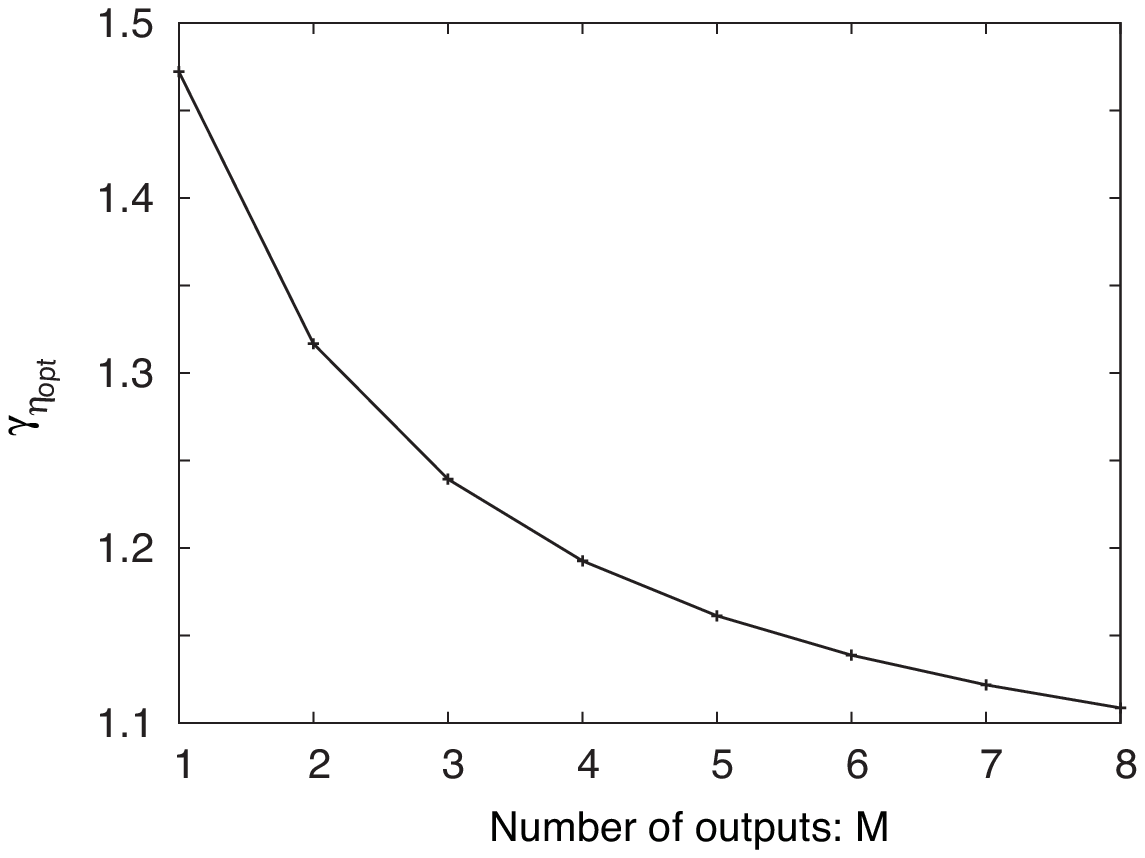}
\caption{\label{pic-opt-eta-gamma}Optimum learning step size $\eta_{opt}$(left) and optimum ratio $\gamma_{\eta_{opt}}$ minimizing the residual error when the optimum learning step size is used (right). }
\end{figure}

\noindent
In Fig. \ref{pic-opt-eta-gamma}, the optimum learning step size $\eta_{opt}$ and the optimum ratio of the baseline noise to perturbation noise, $\gamma_{\eta_{opt}}$, are shown. The optimum ratio $\gamma_{\eta_{opt}}$ minimizes the residual error when the optimum learning step size $\eta_{opt}$ is used. From the figure, the optimum learning step size $\eta_{opt}$ asymptotically approaches zero for many output units. Moreover, the ratio $\gamma_{\eta_{opt}}$ asymptotically approaches one for many outputs when the optimum learning step size $\eta_{opt}$ is used.
 
\begin{figure}[h]
\begin{center}
\includegraphics[width=7cm]{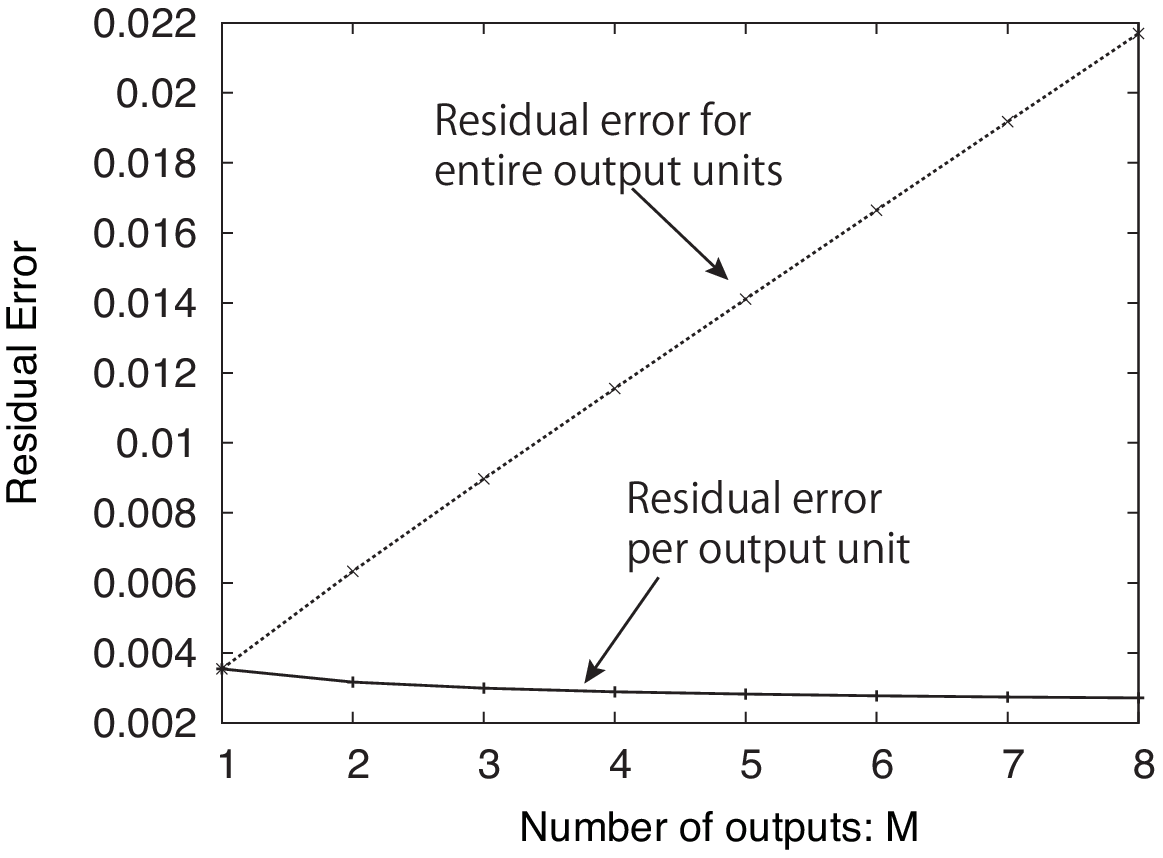}
\end{center}
\caption{\label{opt-eta-gamma-re}Residual error per output unit and that for the entire output units when using the optimum learning step size $\eta_{opt}$ and the optimum ratio $\gamma_{\eta_{opt}}$.}
\end{figure}

Figure \ref{opt-eta-gamma-re} shows numerical calculation of analytical solutions of the temporal development of the residual error (Eq.  (\ref{eta_opt_RE})) when using the optimum learning step size $\eta_{opt}$ (Eq.  (\ref{opt-eta})) and the optimum ratio $\gamma_{\eta_{opt}}$ (Eq.  (\ref{opt-eta-gamma})). The horizontal axis is the number of outputs $M$, and the vertical axis is the residual error. In the figure, the solid line shows the residual error per output unit, and the dotted line shows the residual error of the entire output units. The figure shows that the residual error per output unit decreases as the number of output units increases. This is caused by the smaller optimum learning rate for a larger number of output units. However, the difference in the residual error for the change in the number of output units is small, so the residual error for the entire output units monotonically increases as the number of output units increases. 

Next, we discuss minimizing the residual error. The ratio $\gamma_{opt}$ that minimizes the residual error is given by using Eq. (\ref{lerr}) and $\partial RE/\partial \gamma=0$:

\begin{equation}
\gamma_{opt}=\frac{2-(2+M)\eta-2\sqrt{(M^2+3M+1)\eta^2-2(M+1)\eta+1}}{M\eta}.
\end{equation}

\begin{figure}[h]
\begin{center}
\includegraphics[width=7cm]{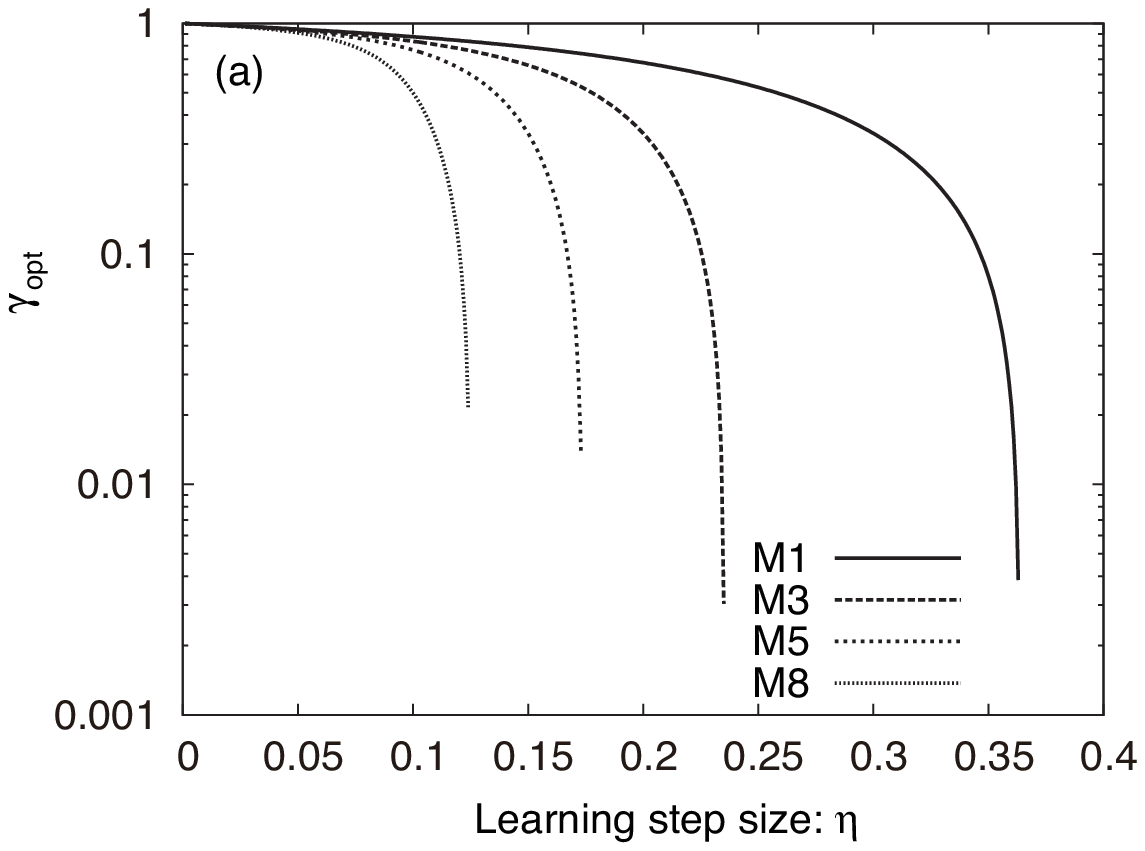}
\includegraphics[width=7cm]{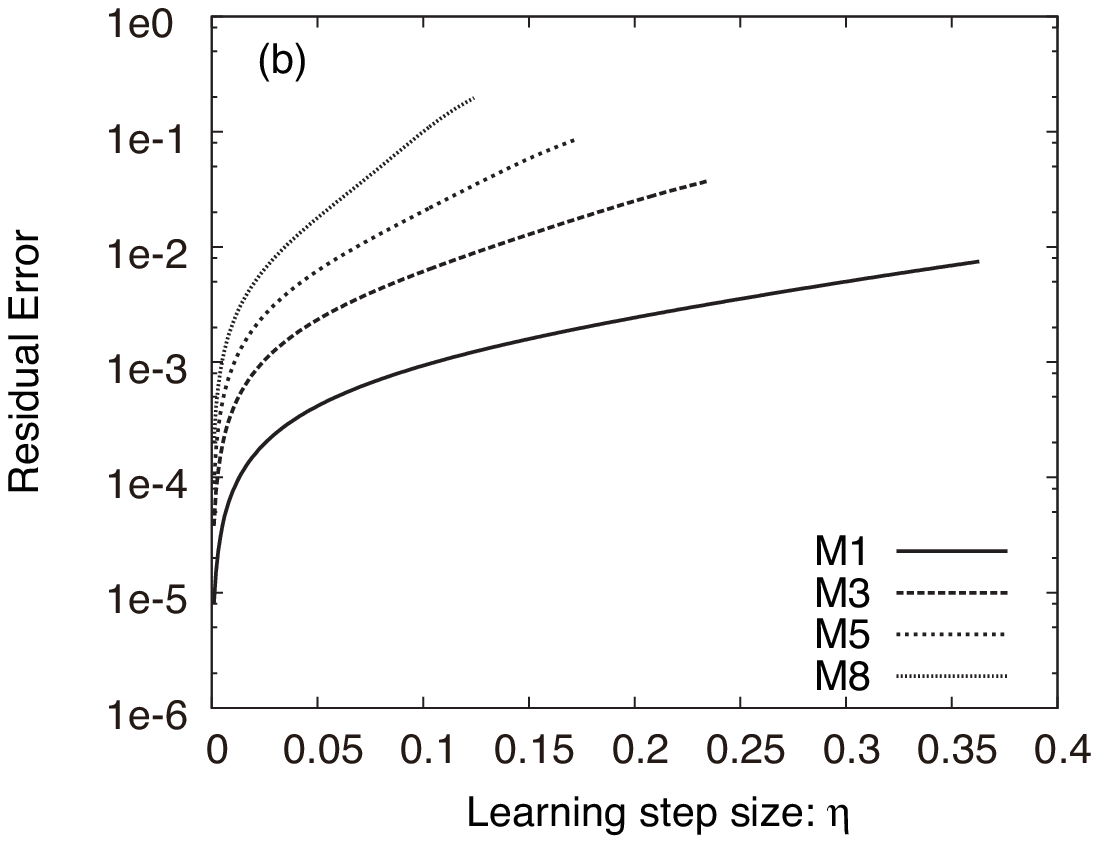}
\end{center}
\caption{\label{opt-gamma} (a) Optimum ratio $\gamma_{opt}$ and (b) residual error using $\gamma_{opt}$ for learning step size.}
\end{figure}

Figure \ref{opt-gamma} shows that the optimum ratio $\gamma_{opt}$ minimizes the residual error and the residual error using $\gamma_{opt}$ for the learning step size. The horizontal axis show the learning step size $\eta$, and the vertical axis of Fig. \ref{opt-gamma}(a) shows the optimum ratio $\gamma_{opt}$, and that of Fig. \ref{opt-gamma}(b) shows the residual error $RE$. From these figures, the optimum ratio $\gamma_{opt}$ converges to one when the learning step size is almost zero. The residual error also converges to zero for this condition. This means that when the learning step size $\eta$ is nearly zero, the residual error becomes the minimum if the variances of the baseline noise $\zeta_k$ and perturbation noise $\xi_k$ are the same. 

\begin{figure}[h]
\begin{center}
\includegraphics[width=8cm]{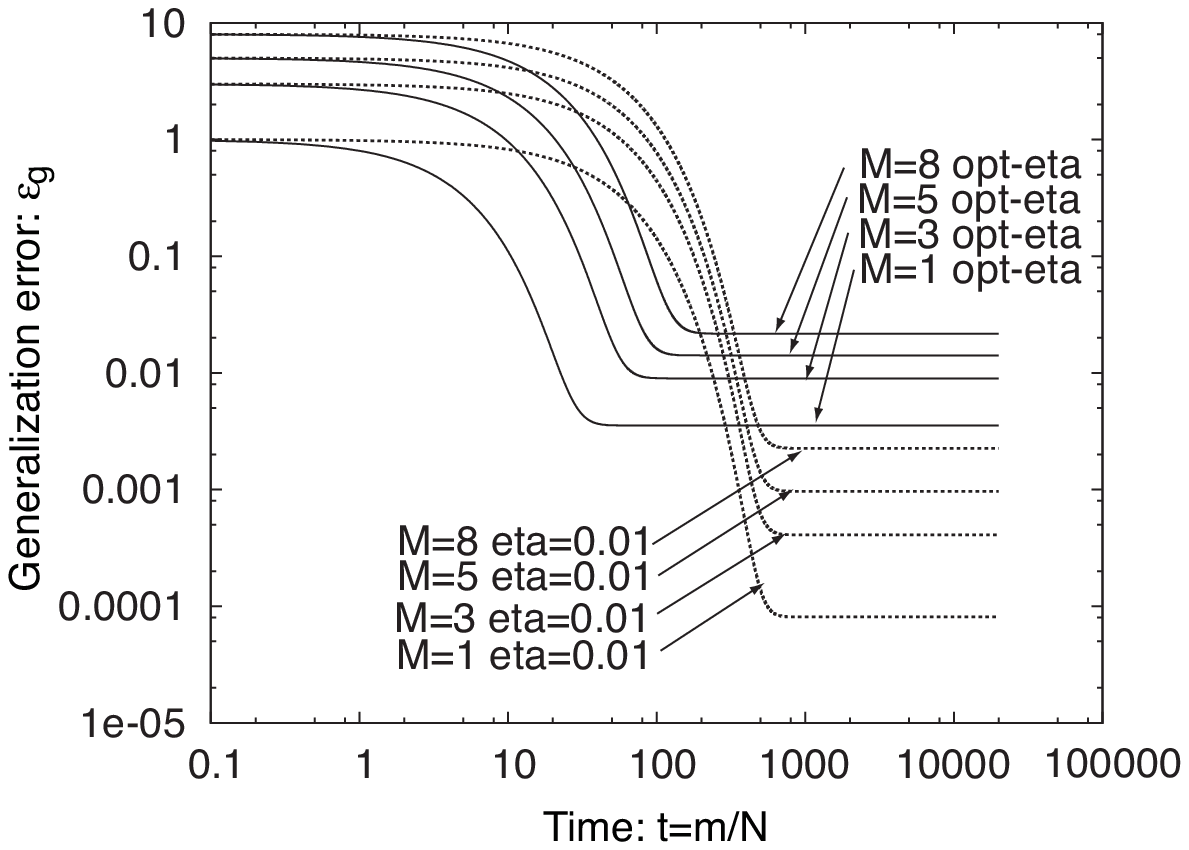}
\end{center}
\caption{\label{small-eta}Comparison of the residual error between using the optimum learning step size $\eta_{opt}$ to maximize the convergence speed and using the small learning step size $\eta=0.01$ to minimize the residual error. }
\end{figure}

\noindent
Taking into account the aforementioned, the learning step size must be almost zero to minimize the residual error. Then, the residual error may not be minimized by the optimum learning step size $\eta_{opt}$. Therefore, we compared the residual error for two cases: setting the learning step size to $\eta_{opt}$ and that of a small learning step size ($\eta=0.01$). Here, we set the optimum ratio of $\gamma_{opt}=1$. In Fig. \ref{small-eta}, we show the results. The horizontal and vertical axes are the same as those in Fig. \ref{fig:eg}. In the figures, the solid lines show the time course of the generalization error using the optimum learning step size $\eta_{opt}$. The lines are also labeled ``M=x opt-eta." Here, x in ``M=x" denotes the number of outputs. The broken lines show the time course of the generalization error using the small learning step size of $\eta=0.01$. These lines are labeled ``M=x eta=0.01." The figures show that the residual error using the small learning step size is smaller than that using the optimum learning step size $\eta_{opt}$.

\begin{figure}[h]
\begin{center}
\begin{minipage}[t]{5cm}
\begin{center}
\includegraphics[width=5cm]{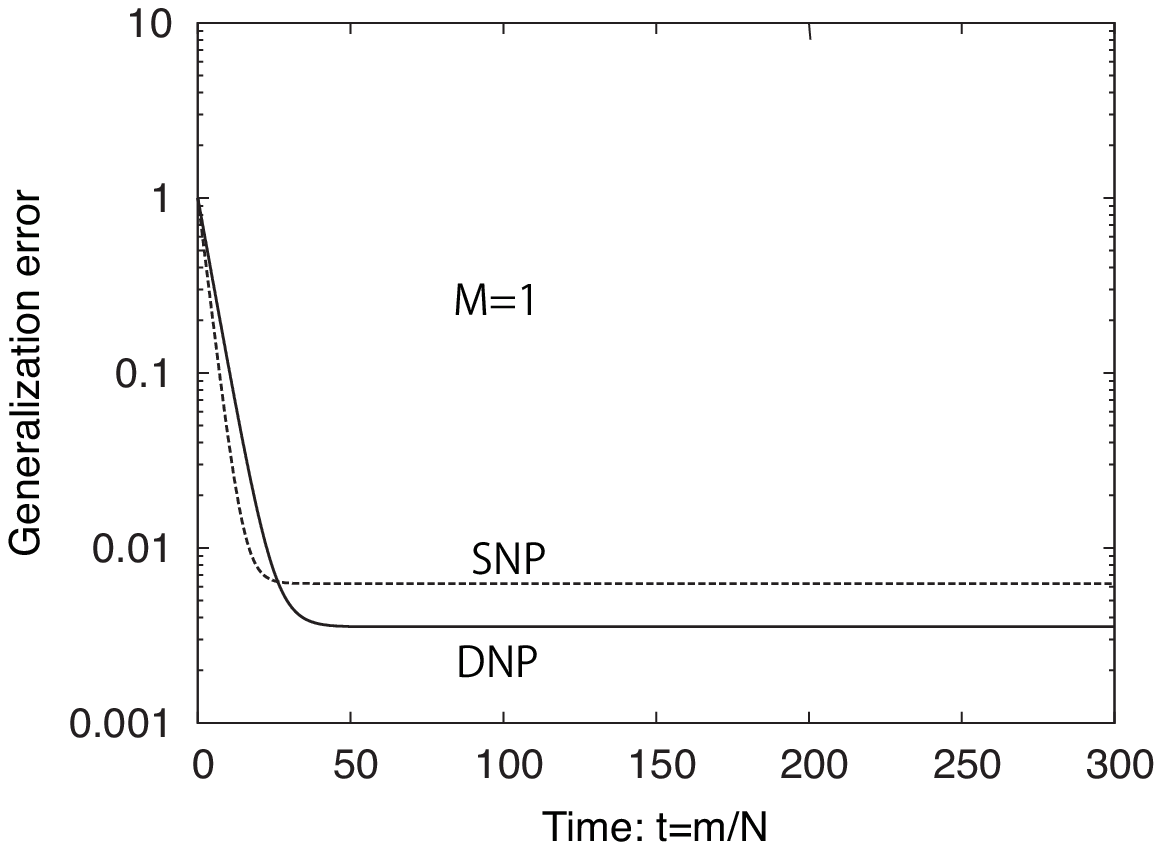}

(a) $M=1$
\end{center}
\end{minipage}
\begin{minipage}[t]{5cm}
\begin{center}
\includegraphics[width=5cm]{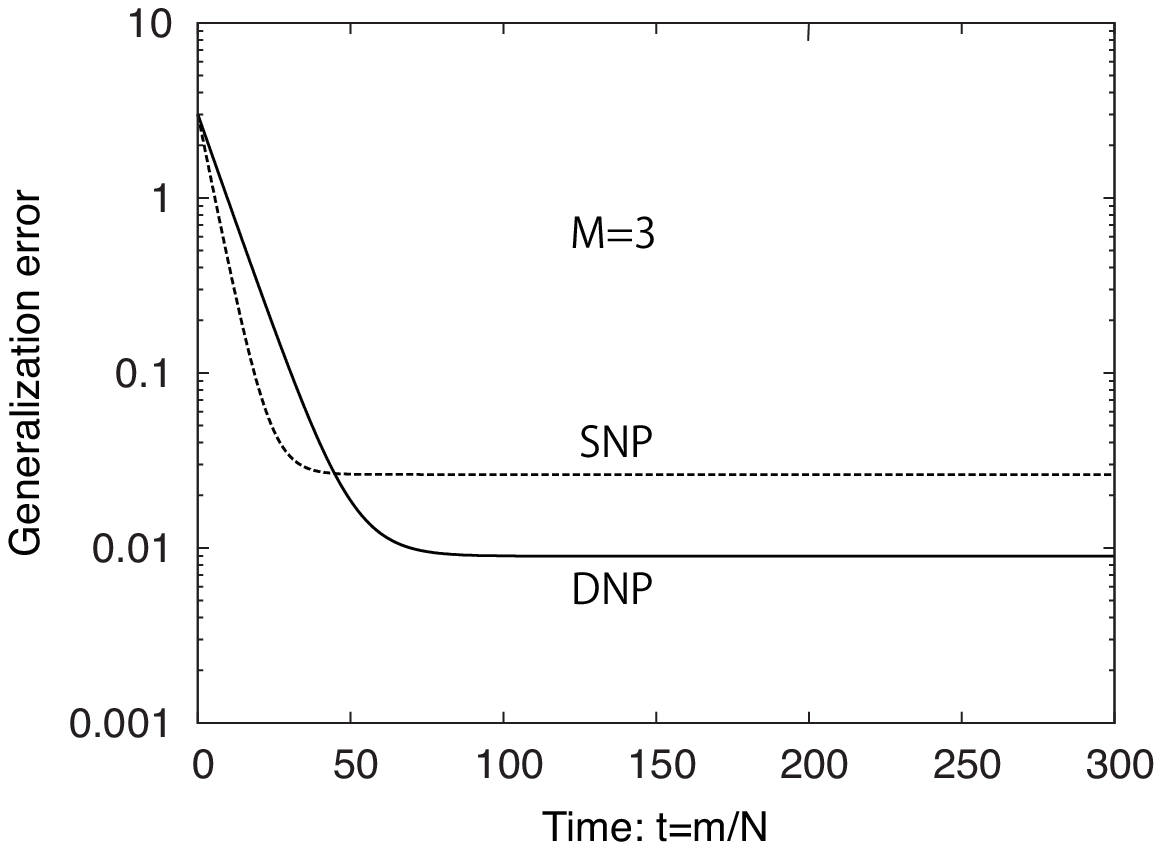}

(b) $M=3$
\end{center}
\end{minipage}

\noindent
\begin{minipage}[t]{5cm}
\begin{center}
\includegraphics[width=5cm]{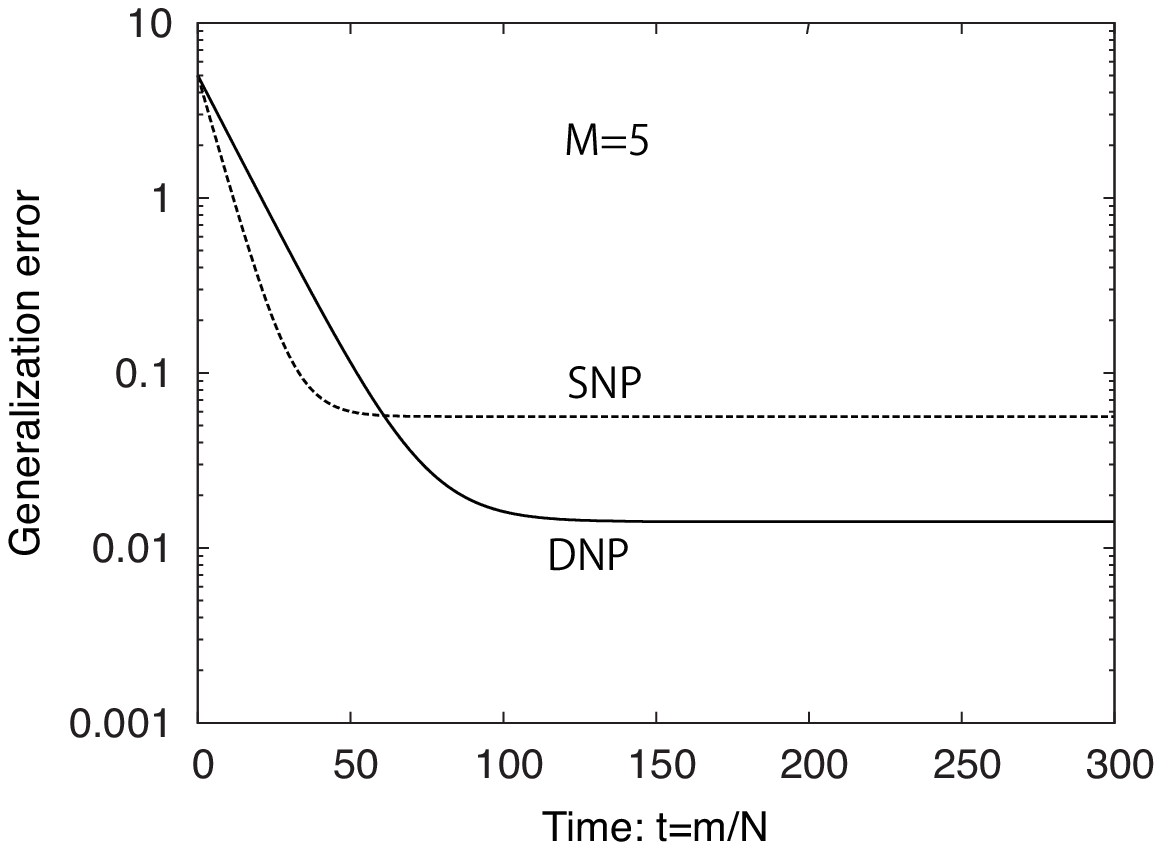}

(c) $M=5$
\end{center}
\end{minipage}
\begin{minipage}[t]{5cm}
\begin{center}
\includegraphics[width=5cm]{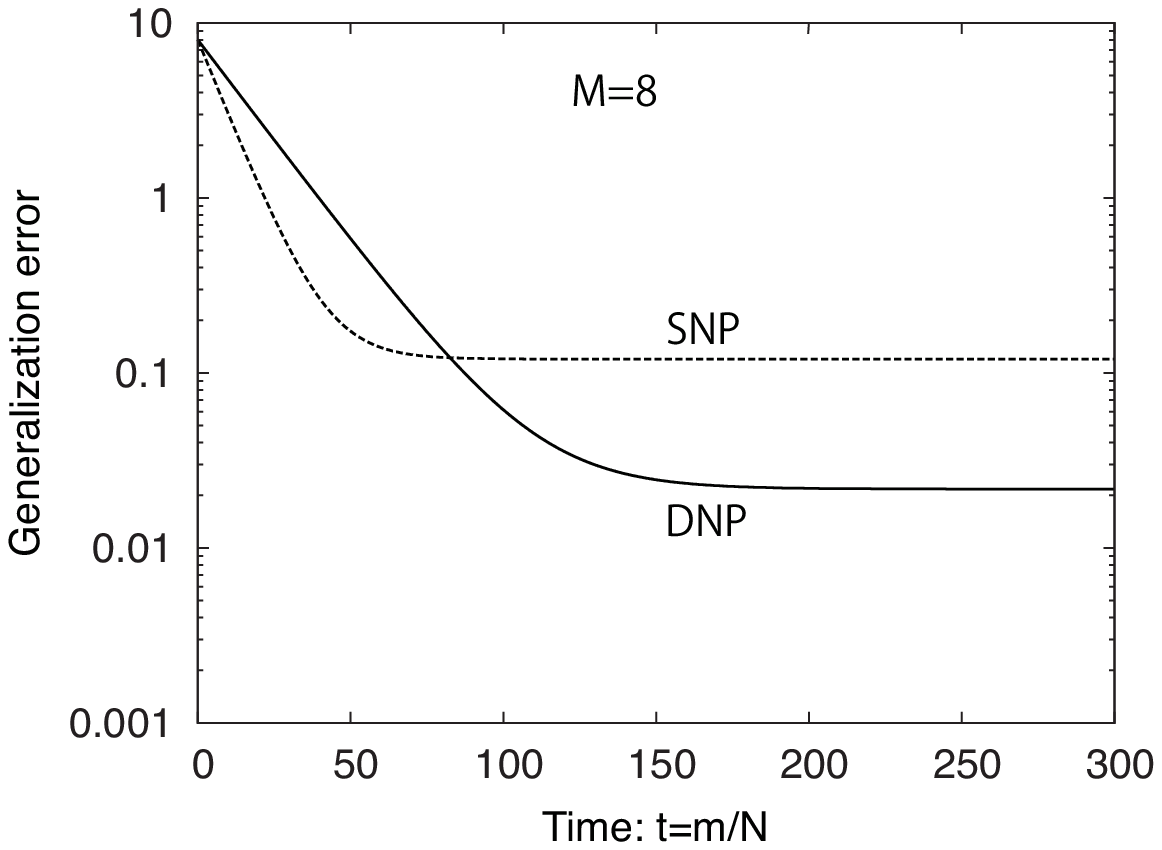}

(d) $M=8$
\end{center}
\end{minipage}
\end{center}
\caption{\label{comp-eg}Comparison of the residual error for SNP or DNP. $M$ is the number of outputs.}
\end{figure}

Finally, we compared the residual error of SNP and that of DNP by performing a numerical calculation of the analytical solutions. The results are shown in Fig. \ref{comp-eg}. The horizontal and vertical axes are the same as those of Fig. \ref{fig:eg}. We used $\sigma_{\xi}^2=0.01$ for DNP and SNP. We used $\eta=0.01$ and $\gamma=1$ for DNP.  $\eta_{opt}$ in Eq. (\ref{opt-eta-1}) is used for SNP. The time course of the generalization for SNP is indicated by the label ``SNP," and that of DNP is indicated by the label ``DNP."  The figures show that the residual error of DNP is smaller than that of SNP for any number of outputs $M$. We suppose that the reason for the smaller residual error of DNP compared with SNP is that DNP is capable of optimizing the ratio $\gamma$ of baseline noise to perturbation noise to minimize the residual error. 

\section{Conclusions}
In this paper, we presented the statistical mechanics of NP  learning, which we built using the baseline noise and perturbation noise. We derived a learning equation by following Cho's formulations and derived coupled differential equations of the order parameters from the learning equation. We solved these equations and obtained analytical solutions. Moreover, we obtained the time course of the generalization error by using the solutions of the order parameters. These results show that using the same variance for the baseline noise and the perturbation noise makes the residual error smallest with a small learning step. The residual error of the proposed method can be smaller than that of NP  learning using only perturbation noise. 

\appendix

\section{NP  learning using only perturbation noise}

We show some analytical results of NP  learning using only perturbation noise\cite{Hara2010}. The differential equations of the order parameters $Q_{kk}$ and $R_{kk}$ are given by 

\begin{align}
\frac{\mbox{d}Q_{kk}}{\mbox{d}t}&=2 \eta (R_{kk}-Q_{kk}) + \eta^2 \Bigl[3(T_{kk}-2R_{kk}+Q_{kk}) \Bigr. \nonumber \\
& +\sum_{k\neq l}^M(T_{ll}-2R_{ll}+Q_{ll}) +\frac{\sigma_{\xi}^2}{4}(M+2)(M+4)\Bigr],\label{dl2dt-1} \\
\frac{\mbox{d}R_{kk}}{\mbox{d}t}&=\eta(T_{kk}-R_{kk}).\label{drdt2-1}
\end{align}

By assuming the symmetry of the evolution equation for updating the weight vector, the analytical solution of the generalization error is given by

\begin{align}
\varepsilon_g&=M\Biggl\{\Bigl(\varepsilon_g^{\ast}(0)-RE\Bigr)e^{-[2\eta-(M+2)\eta^2]t}+RE\Biggr\}, \\
RE&=\frac{(M+2)(M+4)\eta\sigma_{\xi}^2}{8[2-(M+2)\eta]}.
\label{eg-1} 
\end{align}

\noindent
Here, the learning step size that achieves the fastest convergence is given by solving $\frac{d \ \eta(2-(M+2)\eta)t}{d\eta}=0$; then, we obtain the equation

\begin{equation}
\eta_{opt}=\frac{1}{M+2}.\label{opt-eta-1}
\end{equation}

\end{document}